\documentclass[lettersize,journal]{IEEEtran}
\usepackage{amsmath,amsfonts}
\usepackage{algorithmic}
\usepackage{array}
\usepackage[caption=false,font=normalsize,labelfont=sf,textfont=sf]{subfig}
\usepackage{textcomp}
\usepackage{stfloats}
\usepackage{url}
\usepackage{verbatim}
\usepackage{graphicx}
\usepackage{cite}
\usepackage{multirow} 
\usepackage{bbm}
\usepackage{pifont}
\usepackage[ruled,linesnumbered]{algorithm2e}
\usepackage{booktabs}
\begin{document}

\title{Dual Class-Aware Contrastive Federated Semi-Supervised Learning}

%
\author{Qi Guo, Yong Qi, Saiyu Qi, and Di Wu

\thanks{ Qi Guo is with the School of Computer Science and Technology, Xi'an Jiaotong University, Xi'an, Shaanxi 710049, China (e-mail: qiguoqg@outlook.com).}
\thanks{ Yong Qi is with the School of Computer Science and Technology, Xi'an Jiaotong University, Xi'an, Shaanxi 710049, China (e-mail: qiy@xjtu.edu.cn).}
\thanks{ Saiyu Qi is with the School of Computer Science and Technology, Xi'an Jiaotong University, Xi'an, Shaanxi 710049, China (e-mail: saiyu-qi@xjtu.edu.cn).}
\thanks{ Di Wu is with the School of Computer Science and Technology, Xi'an Jiaotong University, Xi'an, Shaanxi 710049, China (e-mail: ddiwu98@163.com).}
\thanks{Corresponding author: Saiyu Qi.}
\thanks{This work has been submitted to the IEEE for possible publication. Copyright may be transferred without notice, after which this version may no longer be accessible.}}


\maketitle

\begin{abstract}
Federated semi-supervised learning (FSSL), facilitates labeled clients and unlabeled clients jointly training a global model without sharing private data. Existing FSSL methods predominantly employ pseudo-labeling and consistency regularization to exploit the knowledge of unlabeled data, achieving notable success in raw data utilization. However, these training processes are hindered by large deviations between uploaded local models of labeled and unlabeled clients, as well as confirmation bias introduced by noisy pseudo-labels, both of which negatively affect the global model's performance. In this paper, we present a novel FSSL method called Dual Class-aware Contrastive Federated Semi-Supervised Learning (DCCFSSL). This method accounts for both the local class-aware distribution of each client's data and the global class-aware distribution of all clients' data within the feature space. By implementing a dual class-aware contrastive module, DCCFSSL establishes a unified training objective for different clients to tackle large deviations and incorporates contrastive information in the feature space to mitigate confirmation bias. Moreover, DCCFSSL introduces an authentication-reweighted aggregation technique to improve the server's aggregation robustness. Our comprehensive experiments show that DCCFSSL outperforms current state-of-the-art methods on three benchmark datasets and surpasses the FedAvg with relabeled unlabeled clients on CIFAR-10, CIFAR-100, and STL-10 datasets. To our knowledge, we are the first to present an FSSL method that utilizes only 10\% labeled clients, while still achieving superior performance compared to standard federated supervised learning, which uses all clients with labeled data.
\end{abstract}

\begin{IEEEkeywords}
Federated Learning, Federated Semi-Supervised Learning, Distributed Learning.
\end{IEEEkeywords}

\section{Introduction}
Federated learning (FL) has emerged as a promising distributed learning technique, enabling isolated clients to collectively train a global model without sharing their private data. Due to the distributed nature of data and the cost of data labeling, distributed data from different clients presents not only statistical heterogeneity but also annotation heterogeneity, meaning that there are unlabeled datasets in addition to labeled datasets. While several FL methods, such as FedAvg\cite{mcmahan2017communication}, Fedprox\cite{li2020federated}, SCAFFOLD\cite{karimireddy2020scaffold}, FedNova\cite{wang2020tackling}, and MOON\cite{li2021model}, have achieved remarkable results by addressing statistical heterogeneity, they require supervised learning (SL) with fully labeled data for each client, limiting their practical application.

To leverage widely-existing unlabeled datasets to further improve FL performance, our focus on federated semi-supervised learning (FSSL) with fully labeled and fully unlabeled clients in this work. Existing FSSL methods mainly employ pseudo-labeling and consistency regularization to utilize knowledge of unlabeled data\cite{yang2021federated,liu2021federated,liang2022rscfed}. For instance, FedMatch\cite{jeong2020federated} considers inter-client consistency loss to encourage consistent outputs from different clients to improve the global model. FedConsist\cite{yang2021federated} aims to minimize the difference between pseudo-labels generated from unlabeled images and predictions from the same unlabeled images after augmentation. RSCFed\cite{liang2022rscfed} that performs random sub-sampling over clients to achieve consensus optimizes the labeled and unlabeled clients by supervised cross-entropy and mean-teacher-based consistency loss. 

However, these methods suffer from two main limitations. One is the large deviation of uploaded local models from labeled clients and unlabeled clients. The other is confirmation bias\cite{arazo2020pseudo} induced by noisy pseudo labels. This large deviation of uploaded local models not only stem from the non-independent and identically distributed distribution (i.e., NonIID) of clients' data but also from the difference in training objective functions employed by labeled and unlabeled clients. Specifically, labeled clients generally use cross-entropy for training on labeled data, approximating the actual data distribution with the predicted data distribution through maximum likelihood estimation. Conversely, unlabeled clients typically adopt pseudo-label-based consistency regularization for training on unlabeled data, ensuring that the learned decision boundary lies within a low-density region and that similar data points yield similar outputs. Due to these divergent training objectives, even datasets without statistical heterogeneity exhibit substantial disparities between uploaded local models from labeled and unlabeled clients. Additionally, confirmation bias emerges from overfitting to self-generated labels, as model-generated pseudo-labels often contain substantial noise due to the model's inherent inaccuracies. Consequently, FSSL performance is substantially compromised by these two limitations. Note that RSCFed\cite{liang2022rscfed} alleviates the deviation to a certain extent through performing random sub-sampling over clients. However, reducing the deviation by aggregation of models without considering the difference of training objectives still suffers from large deviations. Moreover, to alleviate confirmation bias, FedConsist\cite{yang2021federated} employs a high confidence threshold to filter out inaccurately pseudo-labeled data, but relying solely on model output without integrating additional information remains vulnerable to confirmation bias.

To address the aforementioned limitations, we aim to uncover consistent knowledge at a deeper level, transcending the disparities between labeled and unlabeled clients in order to facilitate collaboration. While labeled and unlabeled clients exhibit both statistical and annotation heterogeneity, their data distributions can be viewed as components of the global data distribution. This implies that all clients share an implicit representation space corresponding to the global class distribution, irrespective of their data being labeled or not. In essence, the most valuable common knowledge beyond client differences is the global class distribution shared by all clients. Given that prototypes can serve as a straightforward and effective representation for samples within the same class~\cite{snell2017prototypical,yang2022class,mu2021fedproc}, we introduce class prototype contrastive learning to reduce large deviations by guiding all clients towards a common learning direction and mitigate confirmation bias by incorporating contrastive information. 

\begin{figure}[t]
	\centering
	\includegraphics[width=0.98\linewidth]{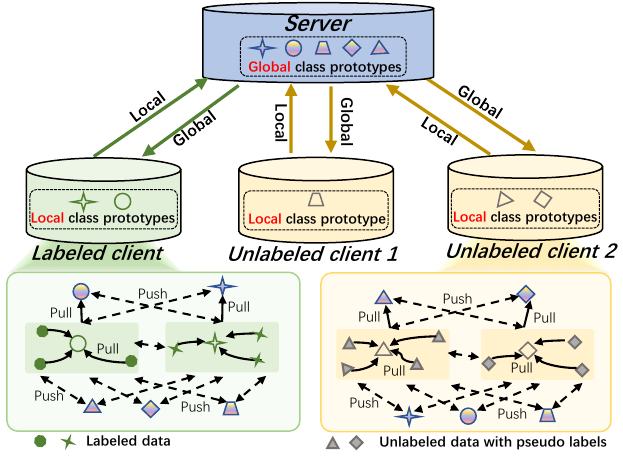}
	\caption{Illustration of the dual class-aware contrastive method with an example of one labeled client and two unlabeled clients. (The local class prototype is an average representation extracted from the same class data in the client, while the global class prototypes are obtained by aggregating the local class prototypes on the server.)}
	\label{intro:fig.illutration}
\end{figure}

Nonetheless, two main challenges persist: one is the mismatch between global and local class prototypes, and the other is the unreliability of class prototypes, particularly for unlabeled clients. Owing to the decentralized nature and statistical heterogeneity of clients' data in FSSL, the class distribution of an individual client's local data (i.e., local class-aware distribution) may diverge from the class distribution of the global data encompassing all clients (i.e., global class-aware distribution). Solely relying on the local class-aware distribution is inadequate to address the challenges posed by NonIID data in the context of data statistical heterogeneity. On the other hand, focusing exclusively on the global class-aware distribution may result in inconsistencies between the global class-aware distribution and each client's potential feature space, leading to significant fluctuations in the global model and impairing the local performance of some clients during the training process. Hence, it is crucial to consider the impact of both global class-aware distribution and local class-aware distribution simultaneously, which entails conducting contrastive learning of global class prototypes and local class prototypes concurrently. Furthermore, due to the inherent inaccuracies in the model's output, the generated prototypes may not be entirely reliable. For unlabeled clients, it is impossible to classify the generated prototypes based on data labels as we do for labeled clients, given the absence of data labels. To make use of the prototypes of unlabeled data, we can label and classify them using pseudo-labels. However, directly applying pseudo-labels to classify the generated prototypes can result in a considerable decline in model performance. This issue arises from the unreliability of pseudo-labels, which leads to the unreliability of the generated prototypes, ultimately causing the model to train in the wrong direction. Consequently, it is essential to enhance the robustness of the class prototypes.

To this end, we propose Dual Class-aware Contrastive Federated Semi-Supervised Learning (DCCFSSL), which mitigates large deviations of uploaded local models from significantly different training goals and alleviates confirmation bias induced by noisy pseudo-labels. On the one hand, to reduce large deviations from different clients, a unified training goal for different clients is introduced through the dual class-aware contrastive module. On the other hand, to alleviate the confirmation bias, the training process of each client is regularized by employing dual class-aware information in the feature space. As depicted in \figurename~\ref{intro:fig.illutration}, DCCFSSL considers both local and global class-aware distributions within the feature space to tackle the mismatch between global class prototypes and local class prototypes. Moreover, we propose the concept of authentication samples to handle the unreliability of class prototypes. Based on authentication samples, we present authentication-reweighted aggregation techniques to bolster the robustness of the global model and class prototypes. Extensive experiments demonstrate that our method surpasses state-of-the-art approaches on three benchmark datasets and outperforms FedAvg with relabeled unlabeled clients on the CIFAR-10, CIFAR-100, and STL10 datasets. These results suggest that DCCFSSL allows FSSL to achieve competitive performance even when compared to standard federated supervised learning using labeled data for all clients. Meanwhile, DCCFSSL not only substantially enhances model performance but also greatly reduces model volatility under both IID and NonIID settings.

Concretely, our contributions and novelty can be summarized as follows:
\begin{itemize}
	\item 
	We propose DCCFSSL, a novel FSSL method designed to leverage unlabeled datasets to further improve FL performance, which tackles the large deviations of uploaded local models from both labeled and unlabeled clients and mitigates confirmation bias induced by noisy pseudo-labels. 
	
	\item 
	We simultaneously consider both local and global class-aware distributions, employing a dual class-aware contrastive module to address the mismatch between global class prototypes and local class prototypes. Additionally, we propose authentication-reweighted aggregation techniques to enhance the robustness of the global model and class prototypes.
	
	\item 
	Comprehensive experiments demonstrate that our DCCFSSL significantly outperforms other state-of-the-art FSSL methods. Notably, DCCFSSL enables FSSL to achieve substantial improvements even when compared to standard federated supervised learning using labeled data for all clients. Furthermore, DCCFSSL greatly enhances the model stability during the training process.
\end{itemize}

\section{Related Works}
\subsection{Federated Learning}
Federated learning (FL)~\cite{mcmahan2017communication} is a rapidly evolving distributed machine learning paradigm that enables clients to collaborate on training using individual local data without compromising their privacy. Numerous studies have been conducted to address various challenges faced by FL, including heterogeneity~\cite{li2019convergence,zhao2018federated}, communication efficiency~\cite{guo2021hybrid,luping2019cmfl}, and robustness~\cite{li2021ditto,lyu2020privacy}. 
Among these, statistical heterogeneity remains a significant obstacle, arising from the differences in clients' data distributions due to their decentralized nature. FedAvg~\cite{mcmahan2017communication}, the most prevalent FL baseline, has demonstrated its effectiveness in mitigating statistical heterogeneity. Subsequent research efforts in this domain can be categorized into two complementary perspectives: client-based training methods and server-based aggregation methods.

\textbf{client-based training methods} emphasize regulating the deviation between local models and the global model in the parameter space to stabilize the local training phase, such as FedProx~\cite{li2020federated}, SCAFFOLD\cite{karimireddy2020scaffold}, MOON~\cite{li2021model}. 
Specifically, FedProx introduces an $l_2$ regularization between global and local model parameters in local objectives to prevent excessive model drift. SCAFFOLD employs variance reduction to correct local model drift during local training. MOON proposes model contrastive federated learning, which adjusts the local training of different clients by exploiting the similarity between model representations. Several other methods also explore client-based training optimizations by using local drift decoupling and standard regularization methods~\cite{gao2022feddc,mendieta2022local}. 

\textbf{server-based aggregation methods} seek to enhance the efficacy of model aggregation, as exemplified by FedNova~\cite{wang2020tackling}, FedMA~\cite{wang2020federated}, Fedbe~\cite{chen2020fedbe}. FedNova introduces a normalized averaging method to eliminate objective inconsistency, while FedMA attempts a layer-wise approach by matching and averaging hidden elements to construct a shared global model. Fedbe improves robustness by sampling higher-quality global models and utilizing Bayesian model ensemble.

Nevertheless, these methods necessitate supervised samples, while we have to consider the utilization of unlabeled data in FSSL. Most existing methods are not directly applicable to the FSSL setting, as the annotation heterogeneity in training optimization between labeled and unlabeled clients can result in uneven model reliability~\cite{liang2022rscfed}.

\subsection{Semi-Supervised Learning}
Pseudo-labeling and consistency regularization form the primary components of semi-supervised learning. Pseudo-labeling methods generate self-predictions for unlabeled data to produce pseudo-labels, allowing the model to self-validate. This self-training process is considered self-confirming. However, due to the risk of overfitting incorrect predictions during training, pseudo-labeling methods are prone to confirmation bias~\cite{arazo2020pseudo}. High-confidence predictions are typically employed to filter noisy unlabeled data~\cite{berthelot2019mixmatch,sohn2020fixmatch}. In contrast, consistency regularization methods~\cite{jeong2019consistency} adhere to the manifold assumption, which posits that different views of the same image should occupy the same position in high-dimensional space, and consequently generate consistent predictions for various image perspectives. Weak and strong augmentations are used to simulate image perturbations and create distinct views of an image~\cite{berthelot2019mixmatch,sohn2020fixmatch}. Building upon these techniques, pseudo-labeling and consistency regularization training~\cite{zhang2021flexmatch,berthelot2019mixmatch,sohn2020fixmatch} have achieved remarkable performance.

Nonetheless, the aforementioned methods primarily concentrate on centralized data, while FSSL addresses the challenges of distributed data under privacy protection for both labeled and unlabeled clients. Instead of focusing on centralized data containing labeled and unlabeled images, this work presents a novel FSSL method that tackles distributed heterogeneous challenges arising from labeled and unlabeled clients.

\subsection{Federated Semi-Supervised Learning}
FSSL scenarios can be categorized into three groups based on the distribution of labeled data among clients and the server~\cite{anonymous2023federated}: (1) Labels-at-Server, where clients possess only unlabeled datasets and the server holds a labeled dataset~\cite{lin2021semifed,he2021ssfl,jeong2020federated}; (2) Labels-at-Clients, where clients own a hybrid dataset consisting of both labeled and unlabeled data~\cite{zhang2021improving,jeong2020federated}; and (3) Labels-at-Partial-Clients, where some clients have labeled datasets while others have unlabeled datasets~\cite{liang2022rscfed,yang2021federated}. FedMatch~\cite{jeong2020federated} employs inter-client consistency loss to apply FSSL in the first two scenarios. RSCFed~\cite{liang2022rscfed} uses distillation on several sub-consensus models and distance-reweighted model aggregation to implement FSSL in the third scenario. FedConsist~\cite{yang2021federated} adopts FSSL in a realistic COVID region segmentation, which falls under the third scenario. In this work, our focus lies on the third scenario, aiming to leverage widely-existing unlabeled datasets to further enhance FL performance.

\section{Methodology}
\label{sec:methodology}
In this section, we first outline the problem setting and introduce relevant notations. Then, we present our Dual Class-aware Contrastive Federated Semi-Supervised Learning (DCCFSSL) approach for FSSL, detailing the local training of clients and the aggregation process of the server. An overview of our DCCFSSL is provided in \figurename~\ref{method:figure.framework}.

\subsection{FSSL Setting}
In this methodology, we consider the FSSL with fully-labeled and fully-unlabeled clients. DCCFSSL involves $n$ labeled clients (denoted as $C_1, ..., C_n$) and $m$ unlabeled clients (denoted as $C_{n+1}, ..., C_{n+m}$). Labeled client $i$ has its own local dataset $\mathcal{D}_i^{l}=\left\{\left(x_j, y_j\right)\right\}_{j=1}^{N^{l}_{i}}$, where $x_j \in \mathbb{R}^P$ is the $P$-dimensional feature vector of a sample, and $y_j \in 1,2, \ldots, K$ represents its label, and $N^{l}_{i}$ is the size of dataset $\mathcal{D}_i^{l}$. Similarly, unlabeled client $i$ also has its own local dataset $\mathcal{D}_i^{u}=\left\{\left(x_j\right)\right\}_{j=1}^{N^{u}_{i}}$, where $x_j \in \mathbb{R}^P$ is a sample without the label, and $N^{u}_{i}$ is the size of dataset $\mathcal{D}_i^{u}$.

\subsection{Framework}

\begin{figure*}[t]
	\centering
	\includegraphics[width=0.98\linewidth]{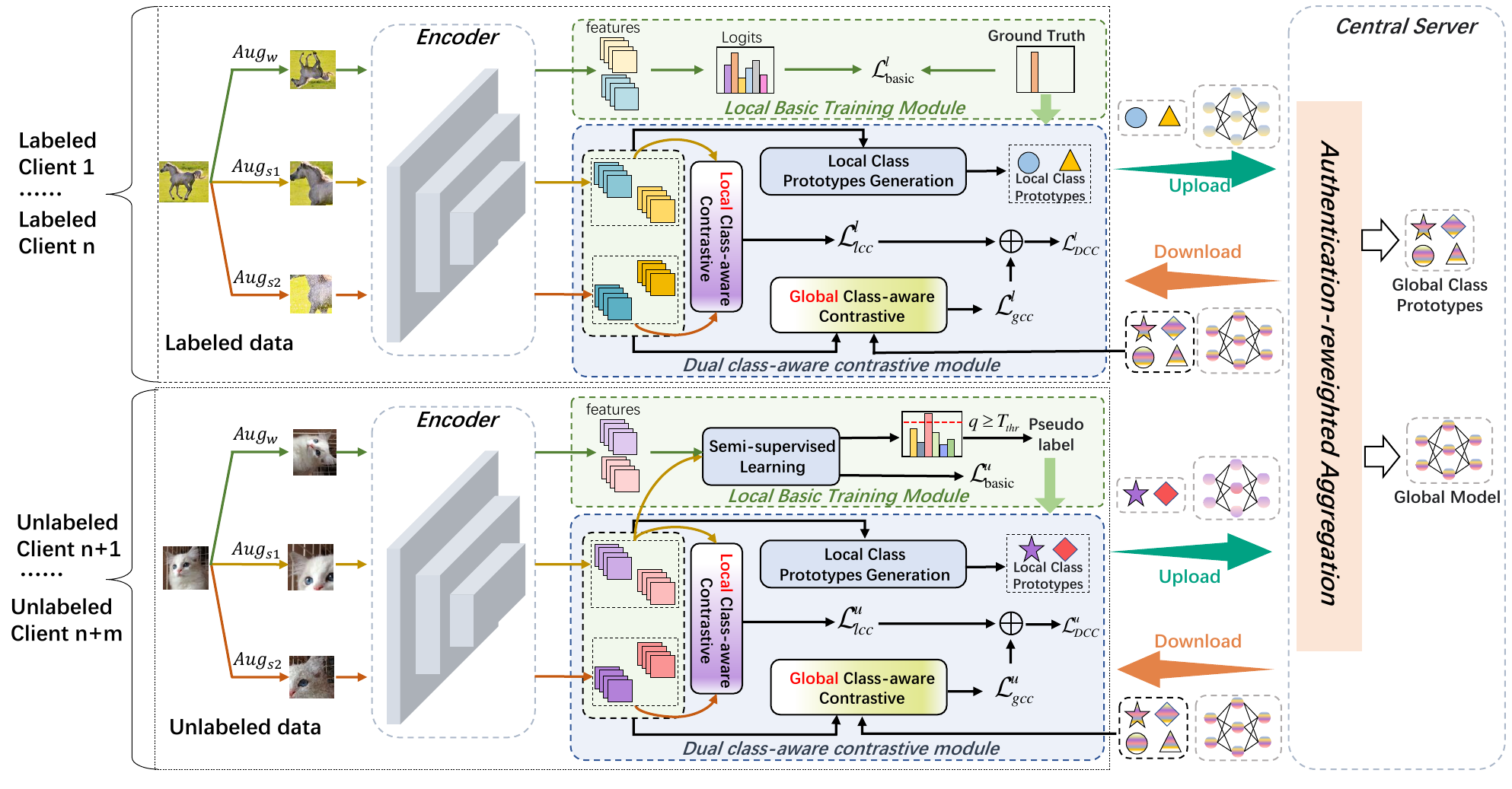}
	\caption{Framework of our proposed DCCFSSL.}
	\label{method:figure.framework}
\end{figure*}

Our proposed DCCFSSL introduces a dual class-aware contrastive module based on the fundamental modules of supervised learning (SL) and semi-supervised learning (SSL), as depicted in~\figurename~\ref{method:figure.framework}. Specifically, in each round, our DCCFSSL conducts the following steps: (1) The client performs local training after receiving a new global model as initialization, then sends the updated model and prototypes back to the server. (2) The server employs authentication-reweighted model aggregation (AMA) and authentication-reweighted prototype aggregation (APA) to obtain the next global model and next global class prototypes on the server; it then randomly samples local clients and sends the current global model and global class prototypes to the selected clients.

\subsection{Local Training}
The local training in FSSL consists of a local basic training module and a dual class-aware contrastive module.

\textbf{Local basic training module} performs the supervised learning of the weak augmentation in labeled clients, and pseudo-label-based consistency learning in unlabeled clients, respectively. For simplicity, we employ the widely used Fixmatch\cite{sohn2020fixmatch} for pseudo-label-based consistency learning in our method.

\textbf{Dual class-aware contrastive module} takes into account both local class-aware distribution and global class-aware distribution. For local class-aware distribution, we regard the representation $z$ from the same category as positive pairs and others as negative for a batch sample. For global class-aware distribution, we treat the global class prototype from the same category as a positive pair and other global class prototypes as negative for a specific sample.

Note that DCCFSSL uses data augmentation and extracts representations of samples as follows\cite{zhang2021flexmatch}.

\textbf{Data augmentation } presents multiple views of a single image. A weak augmentation $Aug_w(\cdot)$ and two strong augmentations $Aug_s(\cdot)$ are used for a labeled or unlabeled image $x_i$. 

\textbf{Encoder} $F(\cdot)$ is used to extract representation $z=F(Aug(x))$ for a given input $x$.

\subsubsection{Labeled clients}
In labeled clients, the basic training module employs a supervised loss $\mathcal{L}_\text{basic}^{l}$ based on $Aug_w(\cdot)$ using the cross-entropy:
\begin{equation}
	\mathcal{L}_\text{basic}^{l}=-y_i \log \left(\hat{y}_i\right),
	\label{eq:method.1}
\end{equation}
where $\hat{y}_i$ is the prediction of local weak augmentation data $Aug_w(x_i)$ from the local model. 

For the dual class-aware contrastive module in labeled clients, we consider the local class-aware contrastive loss $\mathcal{L}_\text{lcc}^{l}$ and the global class-aware contrastive loss $\mathcal{L}_{\text {gcc}}^{l}$.

As in \cite{chen2020simple,he2020momentum}, we randomly sample a minibatch of $N$ images, where encoder $F(\cdot)$ uses two strong augmentations $Aug_s$ of each image $x_i$ to extract features $z$. The local class-aware contrastive loss can be formulated as the following:
\begin{equation}
	\mathcal{L}_\text{lcc}^{l}= -\sum_{i=1}^{2 N} \frac{1}{1+|S(i)|} \sum_{s \in S(i)} \log \frac{  \exp \left(z_i \cdot z_s / \tau\right)}{\sum_{j=1}^{2 N} \mathbbm{1}_{j \neq i} \exp \left(z_i \cdot z_j / \tau\right)},
	\label{eq:method.2}
\end{equation}
where  $\tau$ denotes the temperature parameter and $S(i)$ represents the indices of the views from other images of the same class. $|S(i)|$ denotes its cardinality and $|S(i)|$ + 1 represents all positive pairs.

The global class-aware contrastive loss can be formulated as:
\begin{equation}
	\mathcal{L}_{\text {gcc}}^{l}=-\frac{1}{2N} \sum_{i=1}^{2N} \sum_{j=0}^{C-1} \mathbbm{1}_{y_i=j} \log \frac{\exp{\left(z_i \cdot \bar{z}_j / \tau\right)}}{\sum_{k=0}^{C-1} \exp{\left(z_i \cdot \bar{z}_k / \tau\right)}},
	\label{eq:method.3}
\end{equation}
where $C$ represents the number of classes in total, $\bar{z}_j$ is the global prototype of the $j$-th class sent by the server for local latent feature alignment of samples belonging to class j.

The dual class-aware contrastive loss $\mathcal{L}_\text{DCC}^l$ in labeled client can be expressed as:
\begin{equation}
	\mathcal{L}_\text{DCC}^{l} = \lambda_\text{lcc} \mathcal{L}_\text{lcc}^{l} + \lambda_\text{gcc} \mathcal{L}_{\text {gcc }}^{l},
	\label{eq:method.4}
\end{equation}
where $\lambda_\text{lcc}$ and $\lambda_\text{gcc}$ are coefficient factors used to control the influence of the local class-aware distribution and the global class-aware distribution, respectively.

Finally, the total loss for the labeled client can be formulated as:
\begin{equation}
	\mathcal{L}_\text{total}^{l} = \mathcal{L}_\text{basic}^{l} + \mathcal{L}_\text{DCC}^{l}.
	\label{eq:method.5}
\end{equation}

\subsubsection{Unlabeled clients}
The basic training module in unlabeled clients employs pseudo-label-based consistency regularization as the unsupervised loss $\mathcal{L}_{\text{basic}}^{u}$. Pseudo labels $\hat{q}_i=\arg \max \left(p_i\right)$ on the weak view of image $x_i$ are generated by the model’s prediction $p_i=P_{c l s}\left(A u g_w\left(x_i\right)\right)$ for $\mathcal{L}_{\text{basic}}^{u}$. But consistency training will only retain pseudo labels with high confidence $q \geq T_{\text{thr}}$, where $q=\max (p)$, and $T_{\text{thr}}$ is the threshold to ensure the accuracy of pseudo labels. $\mathcal{L}_{\text{basic}}^{u}$ can be formulated as:
\begin{equation}
	\mathcal{L}_{\text{basic}}^{u}=\frac{1}{N} \sum_{i=1}^{N} \mathbbm{1}\left(\max \left(p_i\right) \geq T_{\text{thr}}\right) H\left(\hat{q}_i, P_{c l s}\left(\operatorname{Aug}_s\left(u_i\right)\right)\right.,
	\label{eq:method.6}
\end{equation}
where $H$ means cross entropy.

Similar to labeled clients, the dual class-aware contrastive module in unlabeled clients contains both the local class-aware contrastive loss and the global class-aware contrastive loss. Due to lacking the label of the sample, we assume the images have a high probability to be reliable representations and should be pulled closer with the same class and pushed away from the other classes. 

The local class-aware contrastive loss in unlabeled clients can be formulated as:
\begin{equation}
	\begin{split}
		\mathcal{L}_\text{lcc}^{u}=& -\sum_{i=1}^{2 N}\mathbbm{1}{\left(\max \left(p_i\right) \geq T_{\text{thr}}\right)} \frac{1}{1+|S(i)|}\sum_{s \in S(i)}\\ &\log \frac{  \exp \left(z_i \cdot z_s / 	\tau\right)}{\sum_{j=1}^{2 N} \mathbbm{1}_{j \neq i} \exp \left(z_i \cdot z_j / \tau\right)},
	\end{split}
	\label{eq:method.7}
\end{equation}
where $S(i)$ represents the indices of the views from other images of the same class with confidence $p>T_{\text {thr}}$. $|S(i)|$ denotes its cardinality and $|S(i)|+ 1$  is all positive pairs.

The global class-aware contrastive loss in unlabeled clients can be formulated as:
\begin{equation}
	\begin{split}
		\mathcal{L}_{\text {gcc}}^{u}=&-\frac{1}{2N} \sum_{i=1}^{2N} \sum_{j=0}^{C-1} \mathbbm{1}{\left(\max \left(p_i\right) \geq T_{\text{thr}} \& y_i=j\right)}  \\ 
		&\log \frac{\exp{\left(z_i \cdot \bar{z}_j / \tau\right)}}{\sum_{k=0}^{C-1} \exp{\left(z_i \cdot \bar{z}_k / \tau\right)}},
	\end{split}
	\label{eq:method.8}
\end{equation}
where $C$ represents the number of classes in total, $\bar{z}_j$ is the global prototype of the $j$-th class.

The dual class-aware contrastive loss $\mathcal{L}_\text{DCC}$ in unlabeled clients can be expressed as:
\begin{equation}
	\mathcal{L}_\text{DCC}^{u} = \lambda_\text{lcc} \mathcal{L}_\text{lcc}^{u} + \lambda_\text{gcc} \mathcal{L}_{\text {gcc }}^{u}.
	\label{eq:method.9}
\end{equation}

Finally, the total loss in unlabeled clients is:
\begin{equation}
	\mathcal{L}_\text{total}^{u} = \mathcal{L}_\text{basic}^{u} + \mathcal{L}_\text{DCC}^{u}.
	\label{eq:method.10}
\end{equation}

\subsection{Authentication-reweighted Prototype Aggregation}
To enhance the robustness of global class prototypes, we propose a new authentication-reweighted prototype aggregation (APA) method. In this context, authentication samples are proposed and defined as correctly classified samples for labeled clients and the samples with high-confidence pseudo labels for unlabeled clients, respectively. Specifically, after completing the local training, each client would use local data to generate local class prototypes $\boldsymbol{o^{i}}=(o_0^{i}, o_1^{i}, ..., o_{C-1}^{i})$ by the new local model. Note that only authentication samples within each client will be employed to produce the corresponding class prototype. Additionally, let $v_{j}^{i}$ represents the number of authentication samples corresponding to the prototype of $j$-th class in $i$-th client. Therefore, the authentication samples number vector is $\boldsymbol{v^{i}}=(v_{0}^{i}, v_{1}^{i},..., v_{C-1}^{i})$ in $i$-th client. Therefore, local class prototype $o_j^{i}$ of $j$-th class in $i$-th client can be formulated as:

\begin{equation}
	o_j^i = \frac{1}{{v_j^i}}\sum\limits_{k \in E_j^i} {z_k},
	\label{eq:method.add.prototype}
\end{equation}
where $E_j^i$ represents the indices of authentication samples of $j$-th class in $i$-th client. 

On the server, we maintain up-to-date local class prototypes $\boldsymbol{o^{i}}$ with the vector $\boldsymbol{v^{i}}$ for each client. In each round, the new $\boldsymbol{o^{i}}$ and $\boldsymbol{v^{i}}$ are sent by $K$ selected clients to the server for updating their prototypes and the number vectors status. Considering the reliability of labeled client's prototypes and the relative uncertainty of the unlabeled client's prototypes, we adaptively expand the authentication samples number vector of labeled clients by a factor $\mu=m/n$ (i.e., the ratio of unlabeled clients to labeled clients) while maintaining that of unlabeled clients unchanged. Then, server will update the next global class prototypes $\boldsymbol{O}=(O_0, O_1, ..., O_{C-1})$ as the following:

\begin{equation}
	{O_j} = \sum\limits_{i = 1}^{n + m} {\frac{{v_j^i}}{{v_j^{total}}}} o_j^i, \\
	v_j^{total} = \sum\limits_{i = 1}^{n + m} {v_j^i}.
	\label{eq:method.12}
\end{equation}

After updating $\boldsymbol{O}$ by APA, the server distributes the new global class prototypes to selected clients for the next round.

\subsection{Authentication-reweighted Model Aggregation}
To further enhance the robustness of the models, we also propose an authentication-reweighted model aggregation (AMA) method. This method dynamically adjusts the aggregation weight of local models according to the number of authentication samples from different clients. 

In each round, the average model can be represented as:

\begin{equation}
	\theta_{a v g}=\sum_{i=1}^K \frac{A_i}{A_{t o t a l}} \theta_i, \\
	A_{t o t a l}=\sum_{i=1}^K A_i.
	\label{eq:method.11}
\end{equation}
where $K$ is the number of selected clients, and $A_{i, i \in \{1, 2, ..., K\}}$ is authentication samples count of client $i$.

The complete description of DCCFSSL is presented in Algorithm~\ref{algorithmDCCFSSL}.

\begin{algorithm}[t]
	\caption{Dual Class-aware Contrastive Federated Semi-Supervised Learning (DCCFSSL)}\label{algorithmDCCFSSL}
	\KwIn{the global model $\boldsymbol{\theta}$, the global class prototypes $\boldsymbol{O}$, the labeled dataset $\mathcal{D}_i^{l}$, the unlabeled dataset $\mathcal{D}_i^{u}$, mini-batch size $|\mathcal{B}|$, maximum training round $R$, and number of subset $K$.}
	\KwOut{the final model $\boldsymbol{\theta}^{R}$}
	\textbf{Server executes:}\\
	initialize $\boldsymbol{\theta}^{0}$, $\boldsymbol{O}^{0}$\\
	\For{$r = 0$ to $R-1$}{
		Randomly select $\left\{C_{i_l}\right\}_{l=1}^K$ from $n+m$ clients\\
		\For{$k \gets i_1$ \textbf{to} $i_K$ in parallel}{
			send the global model $\boldsymbol{\theta}^{r}$ to $C_{k}$\\
			send the global class prototypes $\boldsymbol{O}^{r}$ to $C_{k}$\\
			$\boldsymbol{\theta}^{r+1}_{k}, \boldsymbol{o^{k}}, \boldsymbol{v^{k}} \gets$ \textbf{LocalTraining}$(r,k,\boldsymbol{\theta}^{r}, \boldsymbol{O}^{r})$\\
		}
		$\boldsymbol{\theta}^{r+1}$ $\gets$ \textbf{AMA}$(\boldsymbol{\theta}^{r+1}_{k},\boldsymbol{v^{k}}, k = i_1 to i_K)$ by Eq.\ref{eq:method.11} \\
		$\boldsymbol{O}^{r+1}$ $\gets$ \textbf{APA}$(\boldsymbol{O}^{r}, \boldsymbol{o^{k}}, \boldsymbol{v^{k}}, k = i_1 to i_K)$ by Eq.\ref{eq:method.12}\\
	}
	Return the final model $\boldsymbol{\theta}^{R}$.\\
	\textbf{LocalTraining:}$(r, k, \boldsymbol{\theta}^{r}, \boldsymbol{O}^{r})$\textbf{:}\\
	\For{$\text{epoch} = 1$ to $E$}{
		\For{each batch $\mathcal{B}$}{
			\If{Labeled client is True}{
				compute ${L}_\text{total}^{l} = \mathcal{L}_\text{basic}^{l} + \mathcal{L}_\text{DCC}^{l}$ by Eq.\ref{eq:method.5}\\
				$\boldsymbol{\theta}^{r} \gets \boldsymbol{\theta}^{r} - \eta \nabla {L}_\text{total}^{l}$\\
			}
			\Else{
				compute ${L}_\text{total}^{u} = \mathcal{L}_\text{basic}^{u} + \mathcal{L}_\text{DCC}^{u}$ by Eq.\ref{eq:method.10}\\
				$\boldsymbol{\theta}^{r} \gets \boldsymbol{\theta}^{r} - \eta \nabla {L}_\text{total}^{u}$\\
			}
		}
	}
	$\boldsymbol{\theta}^{r+1}_{k} \gets \boldsymbol{\theta}^{r}$\\
	generate local prototypes $\boldsymbol{o^{k}}=(o_0^{k}, o_1^{k}, ..., o_{C-1}^{k})$  with the vector $\boldsymbol{v^{k}}=(v_{0}^{k}, v_{1}^{k},..., v_{C-1}^{k})$ by $\boldsymbol{\theta}^{r+1}_{k}$\\
	Return $\boldsymbol{\theta}^{r+1}_{k}$, $\boldsymbol{o^{k}}$, and $\boldsymbol{v^{k}}$.
\end{algorithm}

\section{Experiments}
In this section, we first outline the experimental setup in detail. Next, we evaluate the effectiveness of DCCFSSL in comparison to state-of-the-art FSSL methods. Additionally, we conduct ablation experiments to assess the specific impact of various components. We also investigate the influence of the dual class-aware contrastive module on the stability of the training model. Then, we examine the influence of key hyperparameters on the performance of DCCFSSL. Moreover, we analyze the additional communication cost introduced by DCCFSSL. Finally, we investigate the effect of different ratios of labeled and unlabeled clients on DCCFSSL.

\subsection{Experimental Setup} 

\subsubsection{Datasets and Network Architecture} 
We perform experiments on three datasets: CIFAR-10~\cite{krizhevsky2009learning}, CIFAR-100~\cite{krizhevsky2009learning}, and STL-10~\cite{pmlr-v15-coates11a}, implementing both the IID and NonIID data partition settings for all three datasets.

\textbf{CIFAR-10}~\cite{krizhevsky2009learning}, as an image classification dataset, consists of 60,000 images across 10 classes, with 6,000 images per class. Each image has a fixed resolution of 32×32. The training set contains 50,000 images (with 5,000 images per class). This test set contains 10,000 images (1,000 images per class).

\textbf{CIFAR-100}~\cite{krizhevsky2009learning} is also an image classification dataset consisting of 60,000 images in 100 classes, with 600 images for each class. All images have a fixed size of 32x32 pixels. The training set contains 50,000 images, with 500 images per class. This test set contains 10,000 images, with 100 images per class. 

\textbf{STL-10}~\cite{netzer2011reading} is an image recognition dataset, where the labeled examples across 10 classes with 1300 images per class are used in the experiments. Each image has a resolution of 96x96 pixels and is presented in color. The dataset is partitioned such that 80\% of the data from each class is allocated to the training set, while the remaining 20\% is reserved for the test set.

The CIFAR-100 dataset, compared to the CIFAR-10 dataset, maintains an equal total number of training samples while featuring a larger number of classes, consequently increasing the complexity of the task in terms of class diversity. In contrast, the STL-10 dataset preserves the same class count as CIFAR-10 but provides significantly fewer overall training samples, thereby implying a heightened difficulty regarding the volume of training data. According to the work~\cite{zhang2021flexmatch}, we employ the Wide ResNet WRN-16-2~\cite{zagoruyko2016wide} as the backbone network for the experiments on CIFAR-10 and CIFAR-100, while the Wide ResNet WRN-10-2~\cite{zagoruyko2016wide} is used as the backbone network for the experiments on STL-10. Subsequently, a fully connected layer is added for classification purposes. It should be noted that the representation output dimension of the backbone network in this study is set to 128. To guarantee a fair comparison, the same backbone and classification network are utilized for both labeled and unlabeled clients across all methods.

\subsubsection{Federated Learning Setting}
Each dataset's training set is divided among a total of 50 clients. For the IID setting, we use random sampling to generate IID data partitions across the 50 clients. For the NonIID setting, we follow existing methods~\cite{li2021model,wang2020federated} using Dirichlet splitting and adopt a Dirichlet distribution $Dir(\gamma)$ ($\gamma=1$ for all benchmark datasets in this work) to produce NonIID data partition among the 50 clients. After the IID or NonIID data partition in clients, 10\% of the clients are randomly selected as fully labeled clients, while the remaining 90\% are unlabeled clients.

\subsubsection{Baselines}
We categorize six baselines into three groups: (1) The standard federated supervised learning method using only labeled clients: FedAvg-SL-Lower\cite{mcmahan2017communication}. The standard federated supervised learning method using labeled clients and relabeled unlabeled clients: FedAvg-SL-Upper\cite{mcmahan2017communication}. (2) The naive combination of FL with semi-supervised learning, for example, FedAvg-FixMatch\cite{mcmahan2017communication,sohn2020fixmatch} and Fedprox-FixMatch\cite{li2020federated,sohn2020fixmatch}. (3) The state-of-the-art FSSL methods corresponding to this work, e.g., FedConsist\cite{yang2021federated}, RSCFed\cite{liang2022rscfed}. The details of the six baselines are presented as follows.

\textbf{FedAvg-SL-Lower}\cite{mcmahan2017communication}: During the FL training, only labeled clients with labeled data are considered for supervised learning, while unlabeled clients are excluded. Labeled clients use the standard FL method, FedAvg, to conduct the entire learning process. 

\textbf{FedAvg-SL-Upper}\cite{mcmahan2017communication}: All clients participate in FL for supervised learning, where the unlabeled data of unlabeled clients is additionally labeled. Then, labeled clients and relabeled unlabeled clients jointly perform supervised learning using the FedAvg method.

\textbf{FedAvg-FixMatch}\cite{mcmahan2017communication,sohn2020fixmatch}: The naive combination of FedAvg with FixMatch is used to tackle federated semi-supervised learning. FixMatch, the most widely used local semi-supervised method, has achieved excellent performance in local semi-supervised learning. FedAvg-FixMatch is a natural extension method to handle labeled clients and unlabeled clients. 

\textbf{Fedprox-FixMatch}\cite{li2020federated,sohn2020fixmatch}: The naive combination of Fedprox with FixMatch is designed for federated semi-supervised learning. Fedprox is a generalization and re-parametrization of FedAvg that incorporates a proximal term to address the inherent heterogeneity in FL. Fedprox-FixMatch considers both system and statistical heterogeneity while addressing the challenges of labeled and unlabeled clients.

\textbf{FedConsist}\cite{yang2021federated}: FedConsist assumes that a generalizable model should produce the same prediction for original data and slightly perturbed data. Using data augmentation, FedConsist ensures consistent predictions from the global model in unlabeled clients to adjust model weights. Specifically, FedConsist minimizes the difference between pseudo-labels generated from unlabeled images and predictions generated from the same unlabeled images after augmentation.

\textbf{RSCFed}\cite{liang2022rscfed}: RSCFed employs random sub-sampling over local clients to achieve consensus when dealing with large deviations from either labeled or unlabeled clients. Instead of directly aggregating local models, the RSCFed algorithm distills several sub-consensus models by randomly sub-sampling clients and then aggregates the sub-consensus models to form the global model.

\subsubsection{Implementation Details}
We conduct experiments with PyTorch using the SGD optimizer. The learning rates in labeled clients and unlabeled clients are empirically set to 0.01 for all methods. The maximum global round is set to 3000, 4000, and 20000 for CIFAR-10, CIFAR-100, and STL-10, respectively. In all cases, the local training epoch is set to 1, and the number of clients selected in each round is 20.

For our method, the default hyperparameters $\lambda_{\text{lcc}}$ and $\lambda_{\text{gcc}}$ in the dual class-aware contrastive loss are both set to 1. The temperature $\tau$ and the high confidence threshold $T_{\text{thr}}$ are set to 1 and 0.95 by default, respectively. To reduce the computational cost, we exclude unlabeled clients from the first half of the global round, while all clients participate as usual in the second half.

For FedConsist\cite{yang2021federated}, we follow the re-implement setting\cite{liang2022rscfed} that enlarges the total weight of selected labeled clients to about 50\% and makes selected unlabeled clients share the remaining 50\% weight in each global round. 

Note that RSCFed\cite{liang2022rscfed} fails to directly generalize in this practical setting, so we re-implement RSCFed in experimental results. Specifically, considering that there is only one labeled client in the original setting of RSCFed and the weight adjustment proposal\cite{liang2022rscfed}, we select all 5 labeled clients and 5 random unlabeled clients in each round, where the number of sub-sampling and the number of clients in each sub-sampling are 2 and 10, respectively. Then we search for the scaling factor hyperparameter $\beta$ from the set $\{1, 100, 10000\}$, which is used for $L_2$ norm of the model gradient between the local model and temporal averaged model within the subset. According to the experimental results, $\beta=100$ achieves the best performance and is set as the default value. Meanwhile, we try to increase the aggregation weight of labeled clients from the set $\{10\%, 50\%, 70\%, 90\%\}$. Our experiments show that 90\% achieve the best classification accuracy. Therefore, we empirically enlarge the total weight of selected labeled clients to about 90\% and make selected unlabeled clients share the remaining 10\% weight in each global round. 

\subsection{Overall Results} 
\begin{table*}[t]
	\centering
	\caption{Results on CIFAR-10, CIFAR-100, and STL-10 datasets under the IID and NonIID settings (\%). Note that the best and second-best results are marked in bold and underlined, respectively.}
	\resizebox{0.98\linewidth}{!}{
		\begin{tabular}{cccccccccccc}
			\hline
			\multirow{2}{*}{Labeling Strategy} & \multirow{2}{*}{Method} & \multicolumn{2}{c}{Client Number} & \multicolumn{4}{c}{IID} & \multicolumn{4}{c}{NonIID} \\ \cline{3-12} 
			&                  & labeled & unlabeled & Acc   & AUC   & Precision & F1    & Acc   & AUC   & Precision & F1    \\ \hline
			&                  &         &           & \multicolumn{8}{c}{CIFAR-10}                                          \\ \hline
			\multirow{2}{*}{Fully supervised} & FedAvg-SL-Upper\cite{mcmahan2017communication}  & 50      & 0         & \underline{74.98} & \underline{96.43} & \underline{75.81}     & \underline{74.85} & \underline{73.65} & \underline{95.96} & \underline{73.56}     & \underline{73.4}  \\
			& FedAvg-SL-Lower\cite{mcmahan2017communication}  & 5       & 0         & 53.64 & 88.95 & 56.15     & 53.86 & 52.05 & 87.91 & 53.15     & 51.21 \\ \hline
			& FedAvg-FixMatch\cite{mcmahan2017communication,sohn2020fixmatch}  & 5       & 45        & 73.26 & 96.32 & 74.58     & 71.47 & 64.33 & 93.8  & 64.54     & 61.23 \\
			& Fedprox-FixMatch\cite{li2020federated,sohn2020fixmatch} & 5       & 45        & 73.24 & 96.26 & 74.21     & 71.92 & 63.88 & 93.79 & 65.78     & 61.38 \\
			Semi supervised                   & FedConsist\cite{yang2021federated}       & 5       & 45        & 66.72 & 93.89 & 67.72     & 66.34 & 60.3  & 92.3  & 62.85     & 57.26 \\
			& RSCFed\cite{liang2022rscfed}           & 5       & 45        & 53.94 & 88.71 & 55.05     & 53.36 & 48.8  & 86.95 & 50.89     & 48.48 \\
			& DCCFSSL(our)     & 5       & 45        & \textbf{86.62} & \textbf{98.7}  & \textbf{86.53}     & \textbf{86.48} & \textbf{83.17} & \textbf{98.02} & \textbf{83.08}     & \textbf{82.86} \\ \hline
			&                  &         &           & \multicolumn{8}{c}{CIFAR-100}                                         \\ \hline
			\multirow{2}{*}{Fully supervised} & FedAvg-SL-Upper\cite{mcmahan2017communication}  & 50      & 0         & \underline{42.10}  & \underline{94.53} & \underline{43.29}     & \underline{42.16} & \underline{41.13} & \underline{94.29} & \underline{41.08}    &\underline{ 40.00}    \\
			& FedAvg-SL-Lower\cite{mcmahan2017communication}  & 5       & 0         & 15.89 & 81.71 & 17.09     & 15.88 & 15.73 & 81.79 & 15.54     & 14.36 \\ \hline
			& FedAvg-FixMatch\cite{mcmahan2017communication,sohn2020fixmatch}  & 5       & 45        & 27.23 & 91.26 & 29.04     & 24.92 & 24.37 & 90.75 & 25.98     & 20.89 \\
			& Fedprox-FixMatch\cite{li2020federated,sohn2020fixmatch} & 5       & 45        & 27.33 & 91.17 & 29.18     & 24.99 & 24.3  & 90.71 & 25.93     & 20.87 \\
			Semi supervised                   & Fed-Consist\cite{yang2021federated}      & 5       & 45        & 12.44 & 80.81 & 21.98     & 12.69 & 13.2  & 81.88 & 21.50      & 10.66 \\
			& RSCFed\cite{liang2022rscfed}           & 5       & 45        & 17.05 & 82.96 & 17.51     & 15.77 & 15.65 & 82.30  & 16.68     & 13.34 \\
			& DCCFSSL(our)     & 5       & 45        & \textbf{49.35} & \textbf{96.18} & \textbf{50.22}     & \textbf{48.79} & \textbf{45.31} & \textbf{95.86} & \textbf{47.08}     & \textbf{44.22} \\ \hline
			&                  &         &           & \multicolumn{8}{c}{STL10}                                              \\ \hline
			\multirow{2}{*}{Fully supervised} & FedAvg-SL-Upper\cite{mcmahan2017communication}  & 50      & 0         & \underline{65.58} & \underline{93.23} & \underline{66.12}     & \underline{65.65} & \underline{62.96} & \underline{92.63} & \underline{63.67}     & \underline{63.17}  \\
			& FedAvg-SL-Lower\cite{mcmahan2017communication}  & 5       & 0         & 47.81 & 86.81 & 48.15     & 47.40 & 41.35 & 83.38 &  44.92      & 37.85 \\ \hline
			& FedAvg-FixMatch\cite{mcmahan2017communication,sohn2020fixmatch}  & 5       & 45        & 60.85 & 92.30 & 63.34     & 60.85 & 48.88 & 87.91 & 51.81     & 47.13 \\
			& Fedprox-FixMatch\cite{li2020federated,sohn2020fixmatch} & 5       & 45        & 60.73 & 92.43 & 62.34     & 60.38 & 48.27 & 87.58 & 51.73     & 45.88 \\
			Semi supervised                   & FedConsist\cite{yang2021federated}       & 5       & 45        & 51.96 & 87.78 & 53.08      & 51.04 & 43.27 & 84.37 & 46.23     & 40.07 \\
			& RSCFed\cite{liang2022rscfed}           & 5       & 45        & 46.27 & 86.26 & 47.09     & 45.88  & 37.96 & 82.10 & 46.28     & 35.17 \\
			& DCCFSSL(our)     & 5       & 45        & \textbf{75.46} & \textbf{96.31}  & \textbf{75.36}     & \textbf{75.33}  & \textbf{65.88}  & \textbf{93.58} & \textbf{66.07}     & \textbf{65.86} \\ \hline
		\end{tabular}
	}
	\label{exp.table.overall}
\end{table*}

The comparative results on three datasets are presented in \tablename~\ref{exp.table.overall}. We can make the following observations:
\begin{itemize}
	
	\item
	The proposed DCCFSSL outperforms the other five baselines (FedAvg-SL-Lower, FedAvg-FixMatch, Fedprox-FixMatch, FedConsist, and RSCFed) with substantially higher accuracy across all three datasets. This superior performance across diverse data heterogeneity settings, datasets, and various benchmark methods convincingly demonstrates the broad generalizability and efficacy of our proposed DCCFSSL.
	
	\item
	DCCFSSL not only surpasses state-of-the-art methods but also demonstrates substantial performance improvements over FedAvg-SL-Upper in both IID and NonIID settings. For instance, on the CIFAR-10 dataset, DCCFSSL achieves accuracy improvements of 11.64\% and 9.52\% over FedAvg-SL-Upper in IID and NonIID settings, respectively. Likewise, on the CIFAR-100 dataset, DCCFSSL exhibits improvements of 7.25\% and 3.18\% in accuracy over FedAvg-SL-Upper for IID and NonIID settings, respectively. Moreover, on the STL-10 dataset, DCCFSSL attains 9.88\% and 2.92\% accuracy enhancements over FedAvg-SL-Upper for IID and NonIID settings, respectively. These results suggest that DCCFSSL, which can necessitate only 10\% of clients to be labeled, outperforms conventional federated supervised learning methods that require all clients to be labeled. This highlights the potential of DCCFSSL to achieve competitive performance in federated semi-supervised learning relative to standard federated supervised learning utilizing fully labeled data.
	
	\item
	FedAvg-FixMatch and Fedprox-FixMatch exhibit modest performance improvements over the FedAvg-SL-Lower method on the CIFAR-10, CIFAR-100, and STL-10 datasets; however, these enhancements are limited and do not attain the performance level of FedAvg-SL-Upper. Although semi-supervised algorithms like FixMatch exhibit remarkable performance, it is essential to consider deeper collaborative characteristics in FSSL. As a result, we propose the dual class-aware contrastive module to improve FSSL in the feature space, emphasizing both local and global class-aware distributions.
	
	\item
	FedConsist and RSCFed exhibit performance improvements of varying degrees, contingent upon the dataset and settings. For example, FedConsist achieves accuracy improvements of 13.08\% and 8.25\% over FedAvg-SL-Lower on the CIFAR-10 dataset in IID and NonIID settings, respectively. Nonetheless, it fails to improve performance on the more challenging CIFAR-100 dataset. These outcomes suggest that the performance fluctuations of FedConsist and RSCFed are more pronounced across different tasks within practical FSSL settings, indicating that these methods are better tailored to specific scenarios.
	
	\item
	FSSL methods exhibit lower performance improvements in the NonIID setting compared to the IID setting, suggesting that the scarcity of labeled clients exacerbates the impact of data statistical heterogeneity in FSSL. In addition to our proposed DCCFSSL achieving the best performance, the second best performance is consistently achieved by FedAvg-SL-Upper. This underlines the significance of data labels in influencing model performance, while also demonstrating that our proposed approach can effectively leverage unlabeled datasets in the absence of data labels.
	
\end{itemize}

\subsection{Ablation Studies} 
\begin{table*}[t]
	\centering
	\caption{Accuracy (\%) and F1 (\%) on different variants of DCCFSSL on various datasets with the IID and NonIID settings. ``$\Delta$" refers to the change value of the variants compared with DCCFSSL.}
	\resizebox{0.98\linewidth}{!}{
		\begin{tabular}{cccccccccccccccc}
			\hline
			&
			\multicolumn{3}{c}{Components} &
			\multicolumn{4}{c}{CIFAR-10} &
			\multicolumn{4}{c}{CIFAR-100} &
			\multicolumn{4}{c}{STL10} \\ \cline{2-16} 
			&
			\multirow{2}{*}{lcc} &
			\multirow{2}{*}{gcc} &
			\multirow{2}{*}{ARA} &
			\multicolumn{2}{c}{IID} &
			\multicolumn{2}{c}{NonIID} &
			\multicolumn{2}{c}{IID} &
			\multicolumn{2}{c}{NonIID} &
			\multicolumn{2}{c}{IID} &
			\multicolumn{2}{c}{NonIID} \\ \cline{5-16} 
			&
			&
			&
			&
			Acc &
			F1 &
			Acc &
			F1 &
			Acc &
			F1 &
			Acc &
			F1 &
			Acc &
			F1 &
			Acc &
			F1 \\ \hline
			DCCFSSL &
			$\checkmark$ &
			$\checkmark$ &
			$\checkmark$ &
			86.62 &
			86.48 &
			83.17 &
			82.86 &
			50.23 &
			49.79 &
			45.80 &
			45.00 &
			75.46 &
			75.33 &
			65.88 &
			65.86 \\ \hline
			DCCFSSL-lcc &
			\ding{55} &
			$\checkmark$ &
			$\checkmark$ &
			86.38 &
			86.38 &
			82.28 &
			81.91 &
			45.12 &
			44.38 &
			42.33 &
			41.21 &
			72.38 &
			72.49 &
			59.42 &
			58.02 \\
			$\Delta$ &
			&
			&
			&
			$\downarrow$0.24 &
			$\downarrow$0.10 &
			$\downarrow$0.89 &
			$\downarrow$0.95 &
			$\downarrow$5.11 &
			$\downarrow$5.41 &
			$\downarrow$3.47 &
			$\downarrow$3.79 &
			$\downarrow$3.08 &
			$\downarrow$2.84 &
			$\downarrow$6.46 &
			$\downarrow$7.84 \\ \hline
			DCCFSSL-gcc &
			$\checkmark$ &
			\ding{55} &
			$\checkmark$ &
			84.68 &
			84.49 &
			80.57 &
			79.80 &
			45.59 &
			45.29 &
			43.52 &
			42.53 &
			74.50 &
			74.39 &
			65.35 &
			64.55 \\
			$\Delta$ &
			&
			&
			&
			$\downarrow$1.94 &
			$\downarrow$1.99 &
			$\downarrow$2.60 &
			$\downarrow$3.06 &
			$\downarrow$4.64 &
			$\downarrow$4.50 &
			$\downarrow$2.28 &
			$\downarrow$2.47 &
			$\downarrow$0.96 &
			$\downarrow$0.94 &
			$\downarrow$0.53 &
			$\downarrow$1.31 \\ \hline
			DCCFSSL-DCC &
			\ding{55} &
			\ding{55} &
			$\checkmark$ &
			81.96 &
			81.83 &
			79.14 &
			78.39 &
			37.61 &
			36.57 &
			36.20 &
			34.15 &
			69.23 &
			68.99 &
			59.73 &
			58.79 \\
			$\Delta$ &
			&
			&
			&
			$\downarrow$4.66 &
			$\downarrow$4.65 &
			$\downarrow$4.03 &
			$\downarrow$4.47 &
			$\downarrow$12.62 &
			$\downarrow$13.22 &
			$\downarrow$9.60 &
			$\downarrow$10.85 &
			$\downarrow$6.23 &
			$\downarrow$6.34 &
			$\downarrow$6.15 &
			$\downarrow$7.07 \\ \hline
			DCCFSSL-ARA &
			$\checkmark$ &
			$\checkmark$ &
			\ding{55} &
			78.56 &
			78.00 &
			68.95 &
			66.69 &
			36.69 &
			33.58 &
			27.01 &
			21.34 &
			71.77 &
			71.37 &
			62.46 &
			59.07 \\
			$\Delta$ &
			&
			&
			&
			$\downarrow$8.06 &
			$\downarrow$8.48 &
			$\downarrow$14.22 &
			$\downarrow$16.17 &
			$\downarrow$13.54 &
			$\downarrow$16.21 &
			$\downarrow$18.79 &
			$\downarrow$23.66 &
			$\downarrow$3.69 &
			$\downarrow$3.96 &
			$\downarrow$3.42 &
			$\downarrow$6.79 \\ \hline
		\end{tabular}
	}
	\label{exp.table.ablation}
\end{table*}

To gain further insights into DCCFSSL, we conduct ablation studies to assess the effectiveness of its various components. Note that the experimental settings are identical except for the specific variables of interest in each group of experiments. We construct four variants of DCCFSSL as follows:

1) $\text{DCCFSSL}^\text{-lcc}$: This variant removes the local class-aware component from the dual class-aware contrastive module, abandoning the local class-aware contrastive loss in both labeled and unlabeled clients.

2) $\text{DCCFSSL}^\text{-gcc}$: This variant removes the global class-aware component from the dual class-aware contrastive module, abandoning the global class-aware contrastive loss in both labeled and unlabeled clients.

3) $\text{DCCFSSL}^\text{-DCC}$: This variant removes the entire dual class-aware contrastive module, abandoning both local and global class-aware contrastive losses in labeled and unlabeled clients.

4) $\text{DCCFSSL}^\text{-ARA}$: This variant removes the authentication-reweighted aggregation based on authentication samples, abandoning authentication samples and authentication-reweighted aggregation for labeled and unlabeled clients.

As illustrated in \tablename~\ref{exp.table.ablation}, we compare the performance of DCCFSSL and its four variants in the IID and NonIID settings across CIFAR-10, CIFAR-100, and STL-10. It is evident that the four variants, $\text{DCCFSSL}^\text{-lcc}$, $\text{DCCFSSL}^\text{-gcc}$, $\text{DCCFSSL}^\text{-DCC}$, and $\text{DCCFSSL}^\text{-ARA}$, exhibit varying degrees of performance degradation compared to DCCFSSL on all three datasets, thereby demonstrating the effectiveness of these components in DCCFSSL. Specifically, we observe that removing the entire dual class-aware contrastive module or its individual parts results in diminished FSSL performance. This suggests not only that the global class-aware component is vital for capturing interactions, but also that the local class-aware component still contributes to DCCFSSL for a better prediction. Moreover, eliminating the entire dual class-aware contrastive module leads to a substantial performance decline, as evidenced by a 4.66\% and 4.03\% reduction in accuracy for $\text{DCCFSSL}^\text{-DCC}$ on CIFAR-10 under the IID and NonIID settings, respectively. This finding underscores the importance of collaboration between the global and local class-aware components for FSSL performance, fostering representations from distinct isolated perspectives to bolster performance. Furthermore, we observe that, even with the presence of a dual class-aware contrastive module, the model's performance significantly decreases without the authentication-reweighted aggregation based on authentication samples. For instance, $\text{DCCFSSL}^\text{-ARA}$ experiences a 13.54\% and 16.21\% reduction in accuracy compared to DCCFSSL on CIFAR-100 under IID and NonIID settings, respectively.

These ablation studies not only highlight the exceptional effectiveness of the dual class-aware contrastive module for FSSL, but also emphasize the individual contributions of the global and local class-aware components. Additionally, the results underscore the effectiveness of authentication-reweighted aggregation based on authentication samples.

\subsection{Stability of Model Training}

\begin{figure*}[t]
	\centering
	\subfloat[The IID setting]{
		\includegraphics[width=0.45\linewidth]{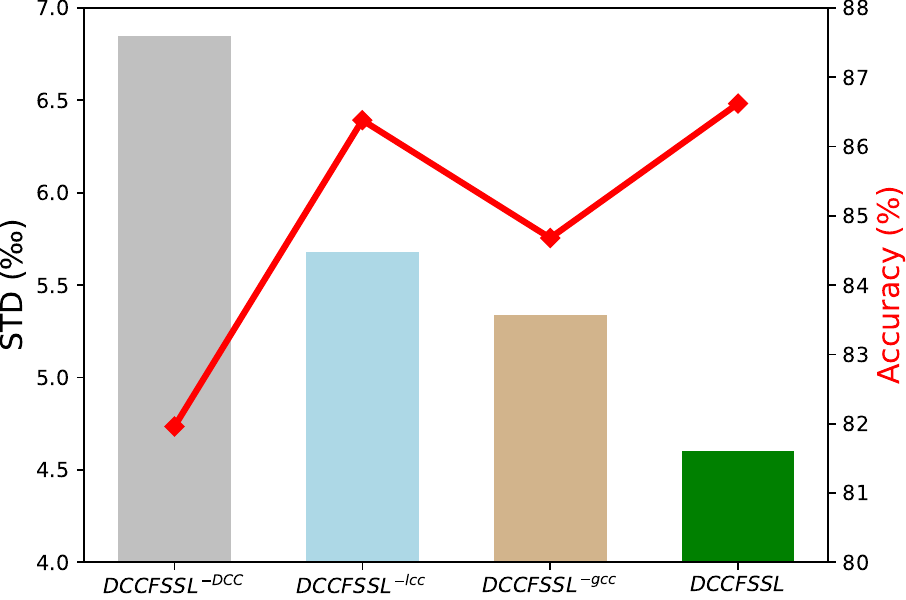}
		\label{fig:short-a}
	}
	\subfloat[The NonIID setting]{
		\includegraphics[width=0.45\linewidth]{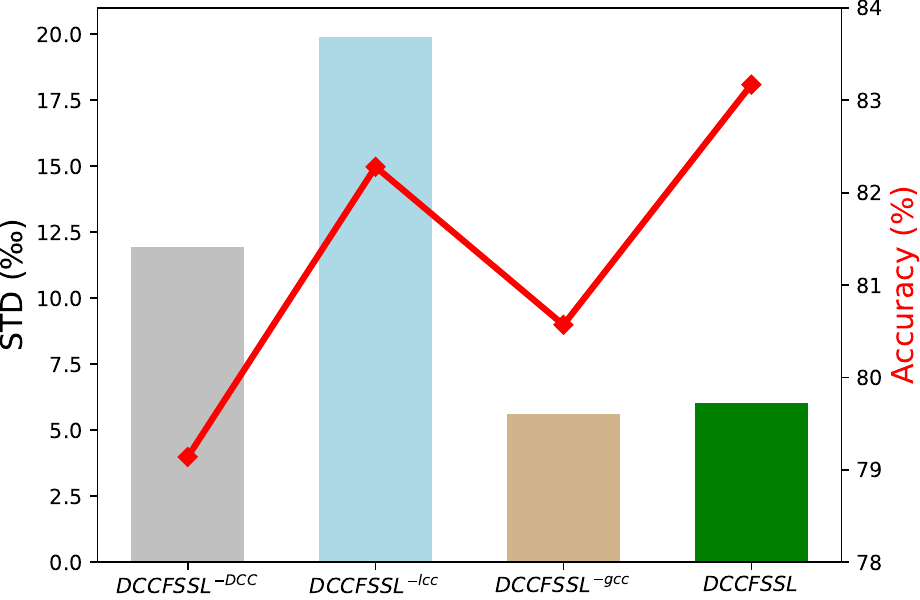}
		\label{fig:short-b}
	}
	\caption{The impact of the dual class-aware contrastive module on the model stability of the  training process on CIFAR-10 (the histogram represents STD, while the line chart displays accuracy.).}
	\label{exp.fig:bar}
\end{figure*}

We further analyze the dual class-aware contrastive module by examining its impact on the stability of the model training process, as illustrated in~\figurename~\ref{exp.fig:bar}. Specifically, we calculate the standard deviation (STD) of the global model's test accuracy for the final 250 global training rounds.

As can be observed from \figurename~\ref{exp.fig:bar}, in the IID setting, the local class-aware and the global class-aware components reduce the global model accuracy fluctuation by 17.08\% and 22.04\%, respectively. Differently, under the NonIID setting, the global class-aware component increases the fluctuation by 66.72\%, while the local class-aware component reduces it by 52.89\%. These findings, together with the ablation studies, indicate that the local class-aware component can reduce the fluctuation of the global model while providing performance improvement to a certain extent. In comparison, the global class-aware component also enhances model performance but increases model volatility in the NonIID setting. In the IID setting, the global class-aware component reduces model volatility due to the homogeneous data distribution among clients. Furthermore, \figurename~\ref{exp.fig:bar} demonstrates that the dual class-aware contrastive module significantly improves model performance while reducing model volatility in both IID and NonIID settings, effectively combining the dual strengths of the local class-aware and global class-aware components at the same time.
\subsection{Impact of Hyperparameter}
\begin{table}[t]
	\centering
	\caption{Accuracy (\%) and F1 (\%) for the hyperparameters  $\lambda_{\text{lcc}}$ and $\lambda_{\text{gcc}}$ of DCCFSSL on CIFAR-10 dataset under both IID and NonIID settings.}
	\resizebox{\columnwidth}{!}{
		\begin{tabular}{ccccccc}
			\toprule
			\multicolumn{2}{c}{Hyperparameter} & Ratio & \multicolumn{2}{c}{IID} & \multicolumn{2}{c}{NonIID} \\ \hline
			$\lambda_{\text{lcc}}$                & $\lambda_{\text{gcc}}$               & $\lambda_{\text{lcc}}$/$\lambda_{\text{gcc}}$   & Acc        & F1         & Acc          & F1          \\ \hline
			0.01             & 0.01            & 1     & 81.72      & 81.41      & 79.34        & 78.81       \\
			0.1              & 0.1             & 1     & 83.59      & 83.49      & 80.66        & 80.28       \\
			1                & 1               & 1     & 86.62      & 86.48      & 83.17        & 82.86       \\
			10               & 10              & 1     & 87.83      & 87.72      & 75.04        & 73.72       \\
			100              & 100             & 1     & 79.08      & 78.24      & 40.24        & 33.57       \\ \hline
			1                & 0.01            & 100   & 84.77      & 84.66      & 80.89        & 80.49       \\
			1                & 0.1             & 10    & 84.83      & 84.72      & 81.51        & 81.02       \\
			1                & 1               & 1     & 86.62      & 86.48      & 83.17        & 82.86       \\
			1                & 10              & 0.1   & 88.81      & 88.74      & 79.67        & 78.52       \\
			1                & 100             & 0.01  & 84.88      & 84.73      & 40.24        & 33.57       \\ \hline
			0.01             & 1               & 0.01  & 86.59      & 86.54      & 83.09        & 82.78       \\
			0.1              & 1               & 0.1   & 86.18      & 86.05      & 82.81        & 82.56       \\
			1                & 1               & 1     & 86.62      & 86.48      & 83.17        & 82.86       \\
			10               & 1               & 10    & 87.07      & 86.93      & 78.71        & 78.27       \\
			100              & 1               & 100   & 86.94      & 86.72      & 78.54        & 77.19       \\ \bottomrule
			
		\end{tabular}
	}
	\label{exp.table.hyper}
\end{table}
To examine the influence of hyperparameters on the performance of our proposed method, we conduct experiments focusing on key hyperparameters $\lambda_{\text{lcc}}$ and $\lambda_{\text{gcc}}$, as demonstrated in \tablename~\ref{exp.table.hyper}.

From the experimental results with the hyperparameter constraint $\lambda_{\text{lcc}}$=$\lambda_{\text{gcc}}$, we observe that maintaining an appropriate ratio between the loss coefficients of the dual class-aware contrastive module and the basic training module is crucial. When this ratio is too small, the dual class-aware contrastive module cannot effectively contribute to performance improvement, limiting its impact. Increasing the ratio moderately can enhance the performance of FSSL. However, when the ratio is too large, it negatively impacts the data representation capabilities of the basic training module, resulting in performance degradation instead of further improvements.

Our experiments with fixed hyperparameters $\lambda_{\text{lcc}}=1$ and $\lambda_{\text{gcc}}=1$ indicate that an overly large or small ratio of $\lambda_{\text{lcc}}$/$\lambda_{\text{gcc}}$ can lead to performance deterioration. It is essential to maintain an appropriate balance between $\lambda_{\text{lcc}}$ and $\lambda_{\text{gcc}}$ to ensure effective collaboration of local and global class-aware components, yielding better performance improvements. For example, under the IID setting, the recommended values for $\lambda_{\text{lcc}}$ and $\lambda_{\text{gcc}}$ are 1 and 10, respectively. Under the NonIID setting, the recommended values for $\lambda_{\text{lcc}}$ and $\lambda_{\text{gcc}}$ are both 1. In the IID setting, where clients have a homogeneous data distribution, increasing the influence of the global class-aware component while maintaining local class-aware component enhances the global collaboration among different clients, thereby improving performance. In the NonIID setting, with heterogeneous data distributions of different clients, excessively increasing the influence of the global class-aware component may amplify the mismatch between individual client's local data distribution and the global data distribution, severely affecting local training and consequently leading to global model performance decline.

When implementing the dual class-aware contrastive module in other works, it is advisable to set both $\lambda_{\text{lcc}}$ and $\lambda_{\text{gcc}}$ to 1 as a simple and effective starting point for further hyperparameter adjustments.

\subsection{Communication Cost}
\begin{table}[t]
	\caption{Parameter quantity of prototype and model in this work. }
	\centering
	\resizebox{0.8\columnwidth}{!}{
		\begin{tabular}{cccc}
			\hline
			Dataset& Prototype & Model  & Ratio  \\ \hline
			CIFAR-10  & 1280      & 692810 & 0.001848 \\ \hline
			CIFAR-100     & 12800     & 704420 & 0.01817  \\ \hline
			STL-10 & 1280      & 692810 & 0.001848 \\ \hline
		\end{tabular}
	}
	\label{exp.table.communication}
\end{table}
In the proposed DCCFSSL method, we upload not only model parameters but also prototype data. We evaluate the added communication cost resulting from the uploading of prototype data. As demonstrated in \tablename~\ref{exp.table.communication}, the incremental communication cost for the CIFAR-10 and STL-10 datasets amounts to a mere 0.1848\% compared to the model's communication cost, while for the CIFAR-100 dataset, the increase is only 1.817\%. This modest rise in communication cost, ranging from 0.1848\% to 1.817\%, is deemed acceptable, particularly in cross-silo scenarios. Moreover, it is important to note that the additional communication cost is solely determined by the dimension of the prototype representation and does not increase proportionally with the growth of the model parameters.

\subsection{Effect of Labeled Client Ratio}
\begin{table}[t]
	\centering
	\caption{The effect of the ratio of labeled clients to unlabeled clients on CIFAR-10 with the IID and NonIID settings.}
	\resizebox{\columnwidth}{!}{
		\begin{tabular}{ccccccc}
			\toprule
			\multirow{2}{*}{Scenarios} & \multicolumn{2}{c}{Client splitting} & \multicolumn{4}{c}{Metrics}                \\ \cline{2-7} 
			& Labeled          & Unlabeled         & Acc(\%) & AUC(\%) & Precision(\%) & F1(\%) \\ \hline
			& 1                & 49                & 75.76   & 96.25   & 76.23         & 75.46  \\
			& 5                & 45                & 86.62   & 98.7    & 86.53         & 86.48  \\
			& 10               & 40                & 88.47   & 99.08   & 88.36         & 88.38  \\
			IID                        & 25               & 25                & 90.47   & 99.38   & 90.48         & 90.42  \\
			& 40               & 10                & 91.68   & 99.49   & 91.73         & 91.70   \\
			& 45               & 5                 & 91.90    & 99.53   & 91.96         & 91.90   \\
			& 50               & 0                 & 91.82   & 99.55   & 91.81         & 91.79  \\ \hline
			& 1                & 49                & 54.84   & 89.31   & 55.86         & 52.48  \\
			& 5                & 45                & 83.17   & 98.02   & 83.08         & 82.86  \\
			& 10               & 40                & 84.47   & 98.47   & 84.42         & 84.30   \\
			NonIID                     & 25               & 25                & 88.64   & 99.19   & 88.63         & 88.55  \\
			& 40               & 10                & 90.51   & 99.44   & 90.66         & 90.51  \\
			& 45               & 5                 & 90.42   & 99.44   & 90.42         & 90.39  \\
			& 50               & 0                 & 90.46   & 99.43   & 90.69         & 90.47  \\ \bottomrule
		\end{tabular}
	}
	\label{exp.table.clientsplitting2}
\end{table}

\begin{figure*}[t]
	\centering
	\subfloat[The IID setting.]{
		\centering
		\includegraphics[width=0.43\linewidth]{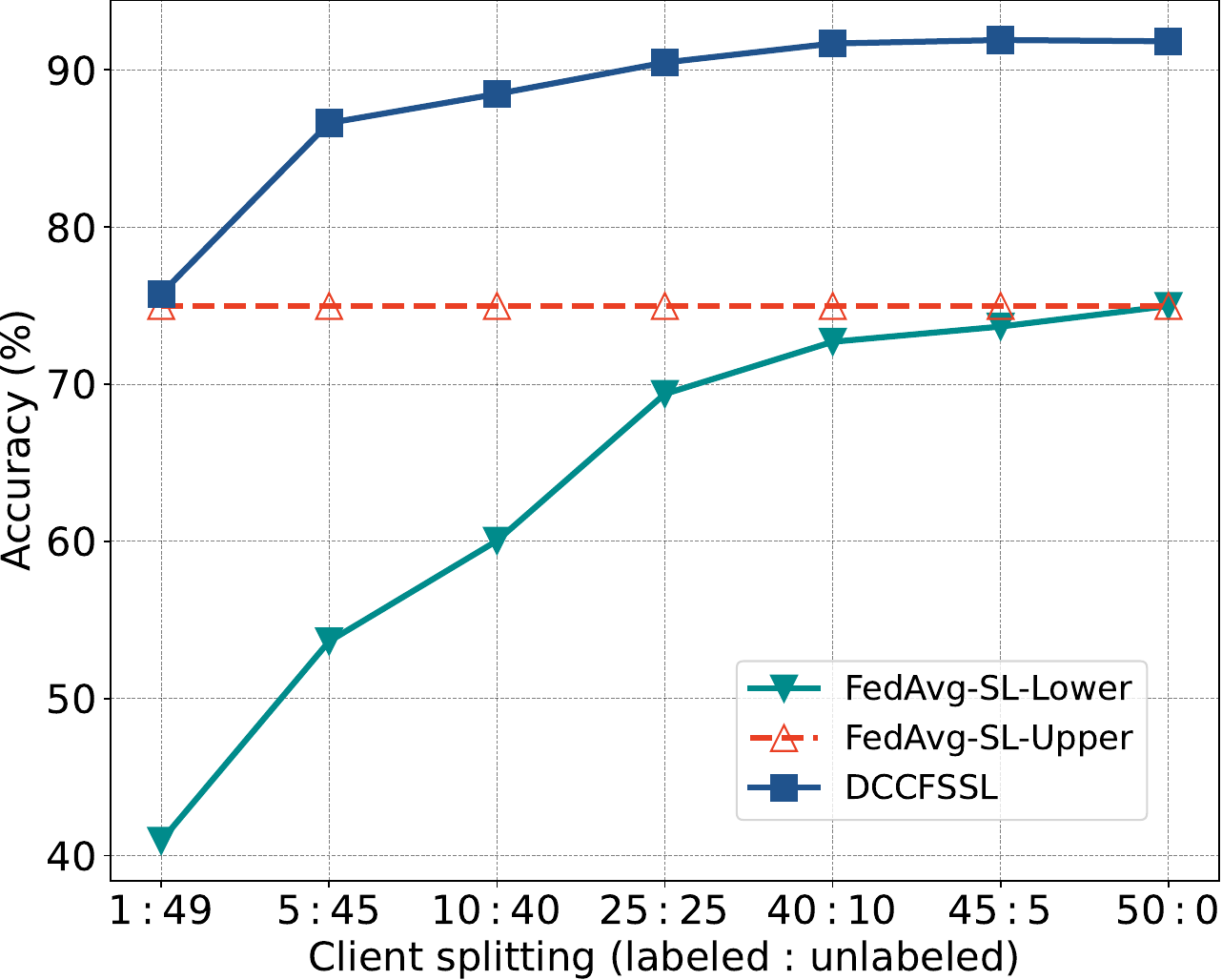}
		\label{fig:short-a}
	}
	\subfloat[The NonIID setting.]{
		\centering
		\includegraphics[width=0.43\linewidth]{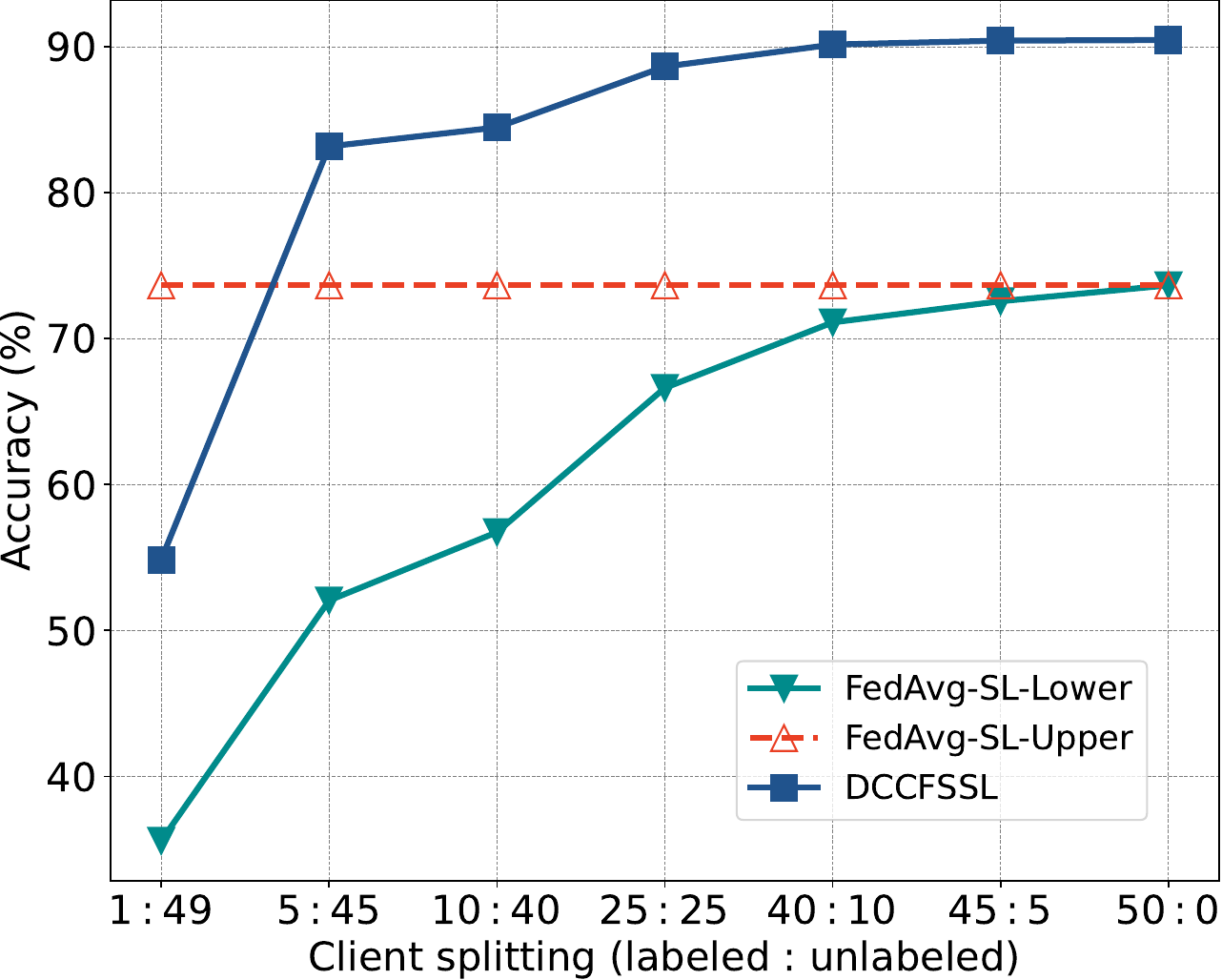}
		\label{fig:short-b}
	}
	\caption{The accuracy of DCCFSSL, FedAvg-SL-Lower, and FedAvg-SL-Upper on different ratios of labeled clients to unlabeled clients.}
	\label{exp.fig:point}
\end{figure*}

In practical scenarios, the ratio of labeled clients to unlabeled clients can vary. Hence, we investigate the impact of different ratios on DCCFSSL performance using the CIFAR-10 dataset. Specifically, we consider seven cases where the ratios of labeled clients to unlabeled clients are 1:49, 5:45, 10:40, 25:25, 40:10, 45:5, and 50:0, respectively. The results of seven cases are as presented in \tablename~\ref{exp.table.clientsplitting2}. We observe that DCCFSSL performance improves as the proportion of labeled clients increases. It should be observed that relying exclusively on labeled clients does not produce the most favorable performance. This phenomenon arises because, when all clients are labeled, a slight decrease in their quantity has a minimal effect on the ultimate accuracy, given the ample availability of labeled data. In contrast, the integration of unlabeled clients contributes consistency information within the context of a high-dimensional space, which leads to a more substantial improvement in accuracy than the reduction associated with diminishing the number of labeled clients. Notably, DCCFSSL achieves 75.76\% and 54.84\% accuracy in the IID and NonIID settings, respectively, even with just one labeled client. This finding suggests that DCCFSSL can perform well even with an extremely limited number of labeled clients.

Moreover, we further compare our method with FedAvg-SL-Lower and FedAvg-SL-Upper, as illustrated in \figurename~\ref{exp.fig:point}. For the IID setting, even with only one labeled client, our method achieves 34.81\%, and 0.78\% improvements over FedAvg-SL-Lower and FedAvg-SL-Upper. For the NonIID setting, our method achieves a 19.26\% improvement over FedAvg-SL-Lower but does not suppress FedAvg-SL-Upper. This result stems from the severely uneven data distribution of a single client, where some classes may have little or no data. However, as the proportion of labeled clients increases, our method's performance significantly improves, surpassing that of FedAvg-SL-Upper in the case of 5:45.

\section{Conclusion}
In this paper, we propose DCCFSSL, a novel federated semi-supervised learning method that leverages unlabeled datasets to further improve FL performance. On the one hand, DCCFSSL builds a common training goal to reduce the large deviation from uploaded local models of labeled clients and unlabeled clients. On the other hand, it introduces dual class-aware contrastive information in the feature space to alleviate confirmation bias. Meanwhile, the proposed dual class-aware contrastive module of DCCFSSL tackles the mismatch between global and local class prototypes by considering both local and global class-aware distributions within the feature space. Moreover, DCCFSSL presents authentication-reweighted aggregation based on authentication samples to improve the robustness of the global model and class prototypes. Extensive experiments demonstrate that DCCFSSL significantly outperforms other state-of-the-art FSSL methods. Meanwhile, DCCFSSL substantially improves the model stability of the training process. Remarkably, even when compared to the standard federated supervised learning with labeled data for all clients, DCCFSSL still achieves 2.92\%$\sim$11.64\% accuracy improvement on CIFAR-10, CIFAR-100, and STL-10 datasets.

\bibliographystyle{IEEEtran}
\bibliography{reference.bib}

\begin{IEEEbiography}[{\includegraphics[width=1in,height=1.25in,clip,keepaspectratio]{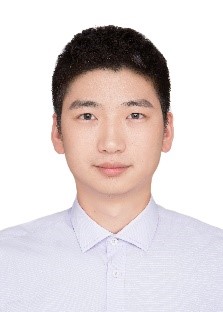}}]{Qi Guo}
	 received the B.S. and M.S. degrees in communication and information systems from Northwestern Polytechnical University, China, in 2017 and 2020. He is currently pursuing the P.H.D. degree in computer science and technology with the Xi’an Jiaotong University, China. His current research interests include federated learning, privacy-preserving machine learning, and artificial intelligence security.
\end{IEEEbiography}
\vspace{-33pt}
\begin{IEEEbiography}[{\includegraphics[width=1in,height=1.25in,clip,keepaspectratio]{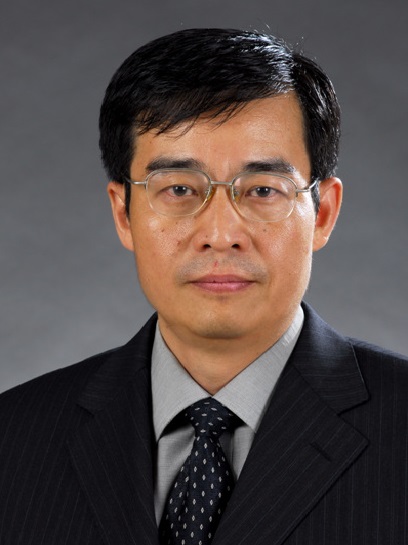}}]{Yong Qi}
	received the PhD degree from Xi’an Jiaotong University, China. He is currently a professor with Xi’an Jiaotong University. His research interests include operating systems, distributed systems, cloud computing, and federated learning.
\end{IEEEbiography}
\vspace{-33pt}
\begin{IEEEbiography}[{\includegraphics[width=1in,height=1.25in,clip,keepaspectratio]{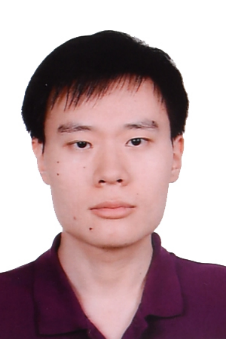}}]{Saiyu Qi}
	received the B.S. degree in computer science and technology from Xi’an Jiaotong University, Xi’an, China, in 2008, and the Ph.D. degree in computer science and engineering from Hong Kong University of Science and Technology, Hong Kong, in 2014. He is currently an Associate Professor with the School of Computer Science and Technology, Xi’an JiaoTong University, China. His research interests include applied cryptography, cloud security, distributed systems, and pervasive computing.
\end{IEEEbiography}
\vspace{-33pt}
\begin{IEEEbiography}[{\includegraphics[width=1in,height=1.25in,clip,keepaspectratio]{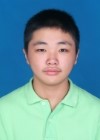}}]{Di Wu}
	received the B.S. degree in Computer Science from Xi’an Jiaotong University, China, in 2019. He is currently pursuing the P.H.D. degree in Computer Science with the Xi’an Jiaotong University, China. His current research interests include privacy-preserving machine learning, federated learning especially on Byzantine-robust Federated Learning.
\end{IEEEbiography}

\vfill

\end{document}